\documentclass[10pt,twocolumn,letterpaper]{article}
\usepackage[pagenumbers]{cvpr} % arxiv
\usepackage{graphicx}
\usepackage{amsmath}
\usepackage{amssymb}
\usepackage[pagebackref,breaklinks,colorlinks]{hyperref}
\usepackage{multirow}
\usepackage{makecell}
\usepackage{wrapfig,boxedminipage,lipsum}
\usepackage[export]{adjustbox}
\usepackage{graphicx}
\usepackage{caption}
\usepackage{balance}

\def\cvprPaperID{24}
\def\confName{CVPR}
\def\confYear{2022}

\newcommand{\Sref}[1]{Section \ref{#1}}
\newcommand{\Tref}[1]{Table \ref{#1}}
\newcommand{\Fref}[1]{Figure \ref{#1}}

\newcommand{\customfootnotetext}[2]{
    {\renewcommand{\thefootnote}{#1}\footnotetext[0]{#2}}
}

\newcommand{\figOverviewFramework}[1]{ 
\begin{figure*}[t]
\centering
\includegraphics[width=\linewidth]{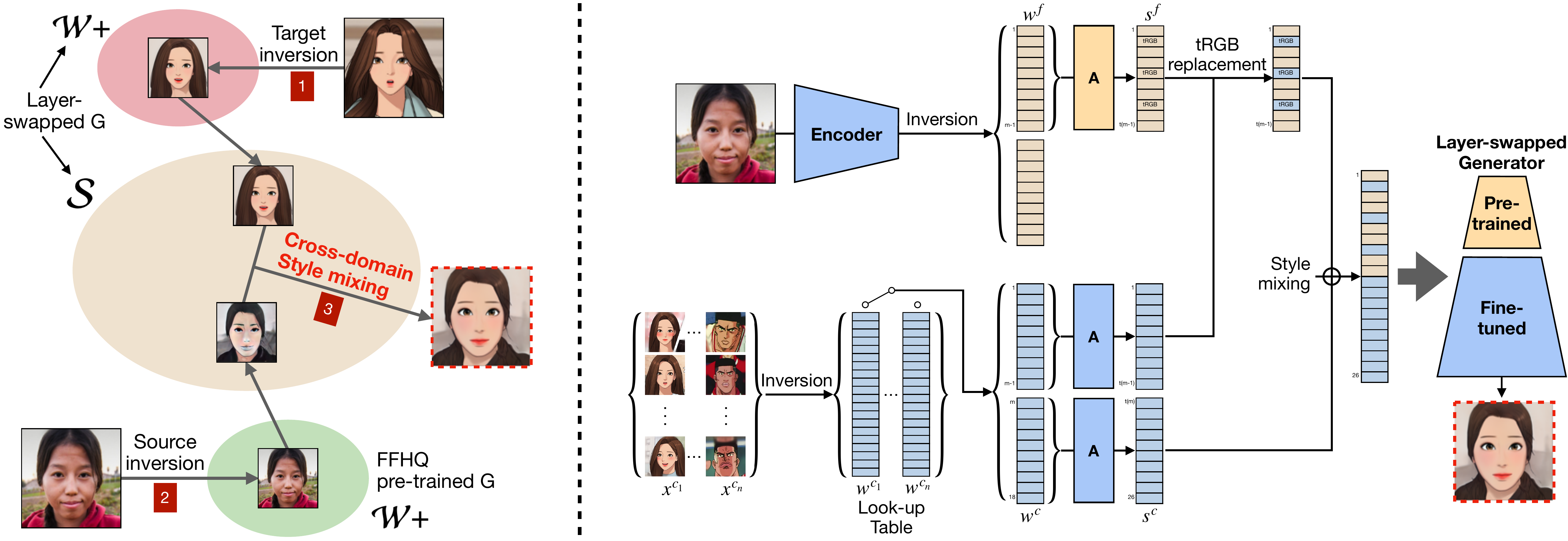}
\caption{Left: Illustration of Cross-domain Style mixing. The latent codes from different domains are style mixed in the same latent space. Right: Model overview. It performs tRGB replacement and style mixing in StyleSpace. We invert the source and target domain images using different inversion methods, and for the latter, all latent codes are stored in a lookup table in advance for inference.}
\label{#1}
\end{figure*}
}

\newcommand{\figCompGenerator}[1]{
\begin{figure}[t]
\centering
\newcommand{\h}{25.5mm}
\newcommand{\himg}{1.0mm}
\makebox[\h][c]{(a)}\hspace{\himg}
\makebox[\h][c]{(b)}\hspace{\himg}
\makebox[\h][c]{(c)}\hspace{\himg}
\\
\includegraphics[width=\h]{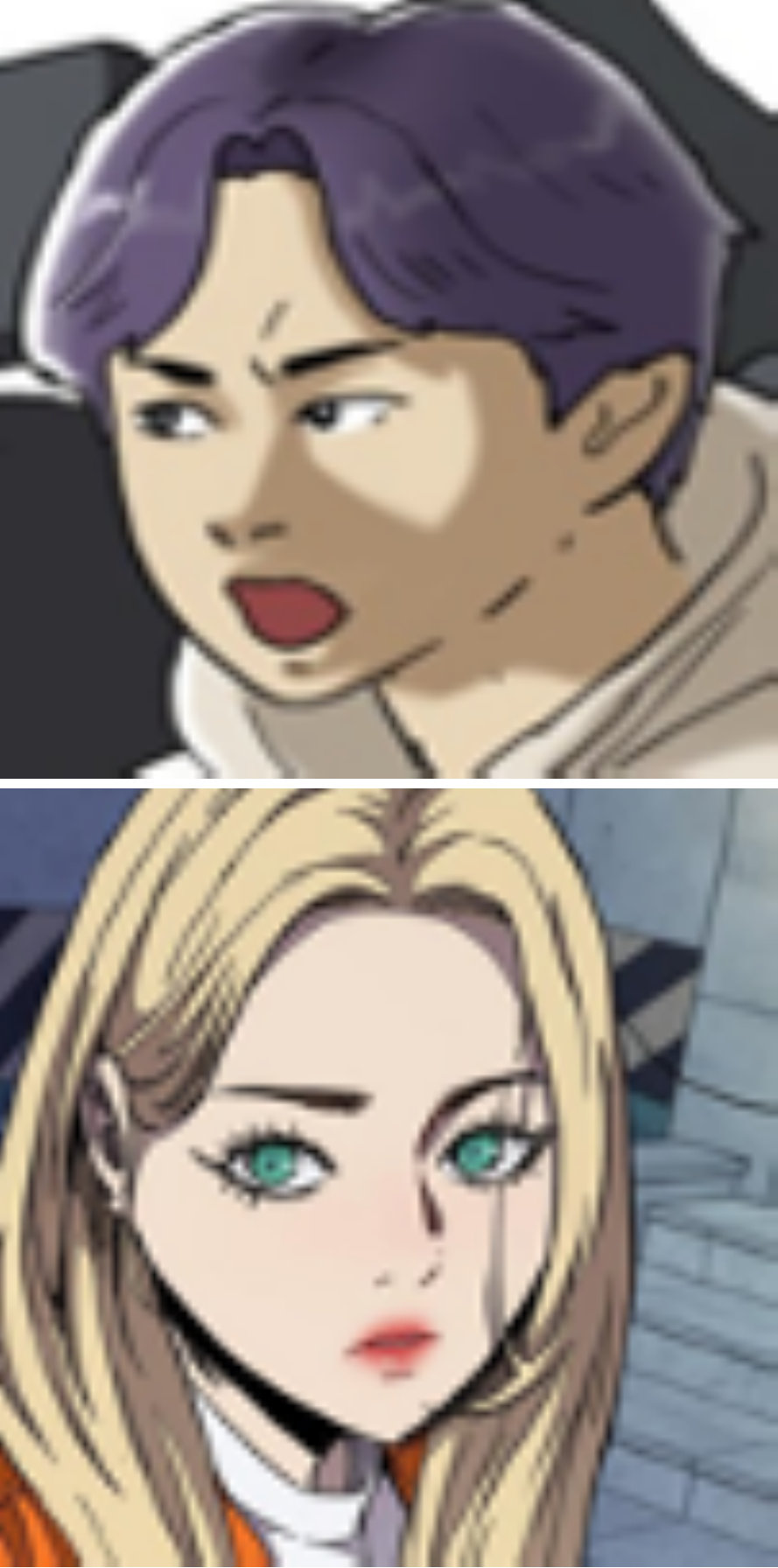}\hspace{\himg}
\includegraphics[width=\h]{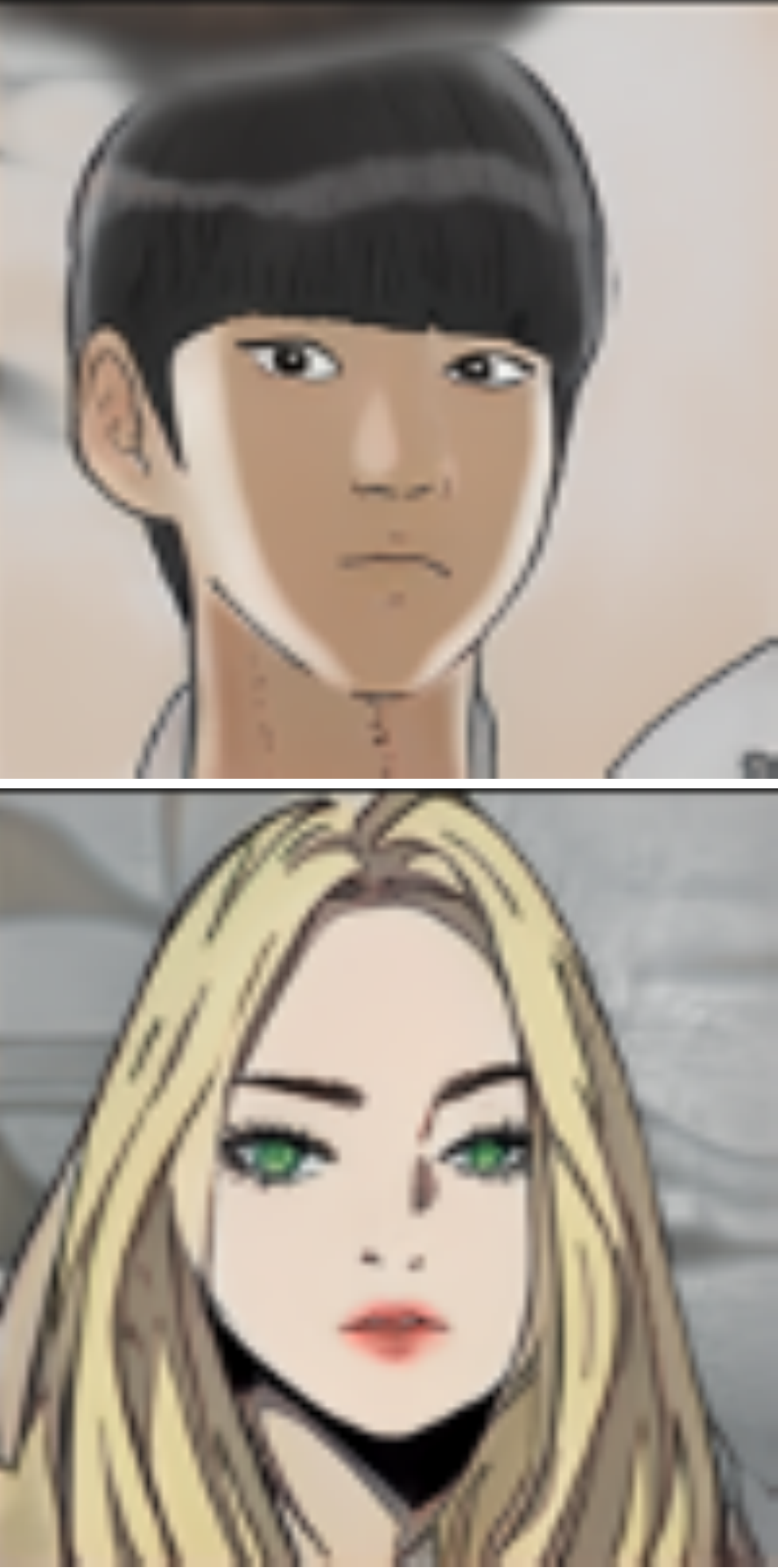}\hspace{\himg}
\includegraphics[width=\h]{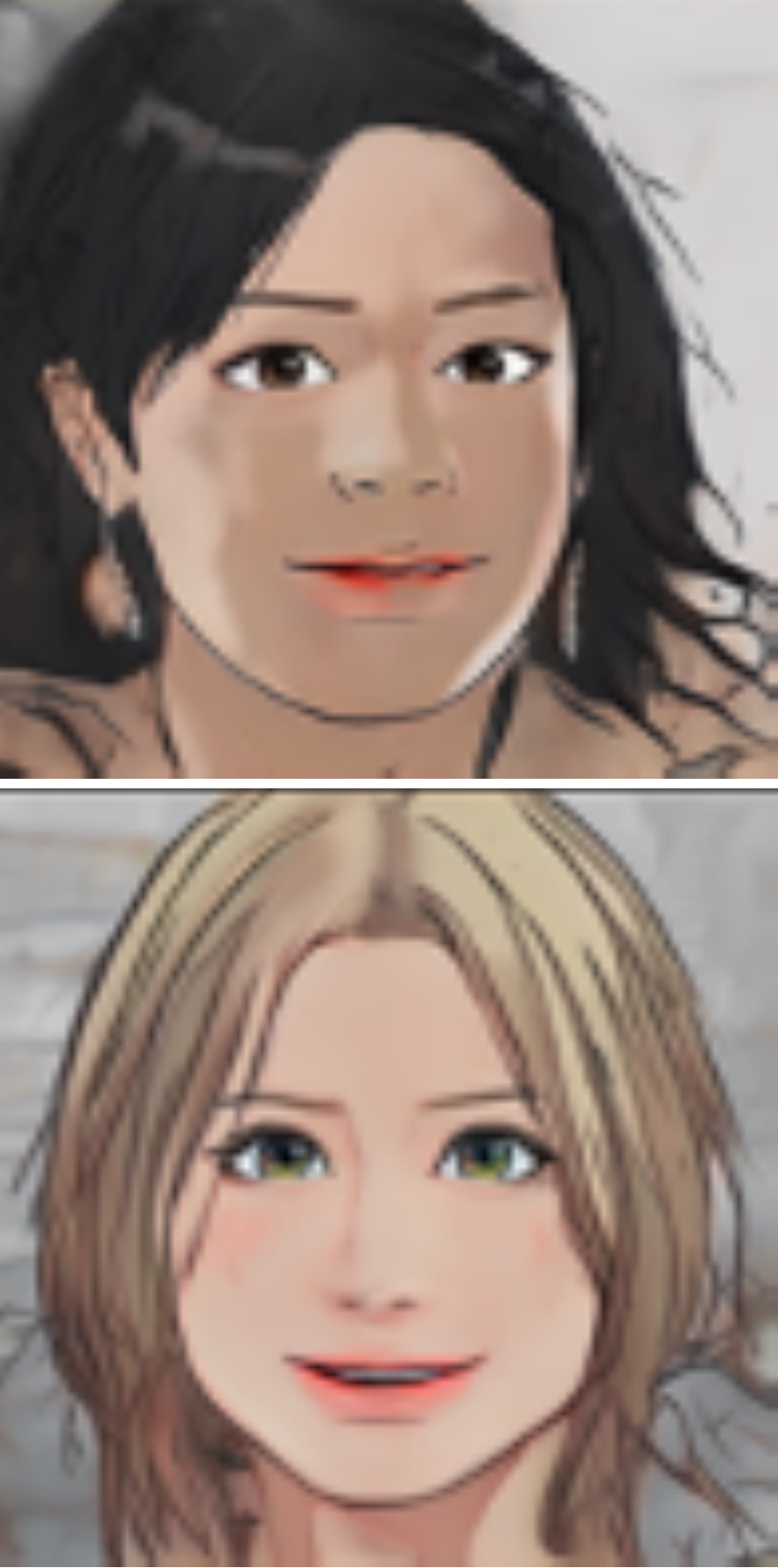}
\caption{Limitation of the layer swapping method. (a) Examples of the cartoon domain dataset. (b) Images generated by the fine-tuned generator. (c) Images generated by the layer-swapped generator. The layer swapping method loses important character-wise cartoon features and instead produces ``mixed'' faces of the natural and cartoon domains.}
\label{#1}
\end{figure}
}

\newcommand{\figCompColor}[1]{
\begin{figure}[t]
\newcommand{\h}{26mm}
\newcommand{\hh}{84mm}
\newcommand{\himg}{2.0mm}
\includegraphics[width=\hh]{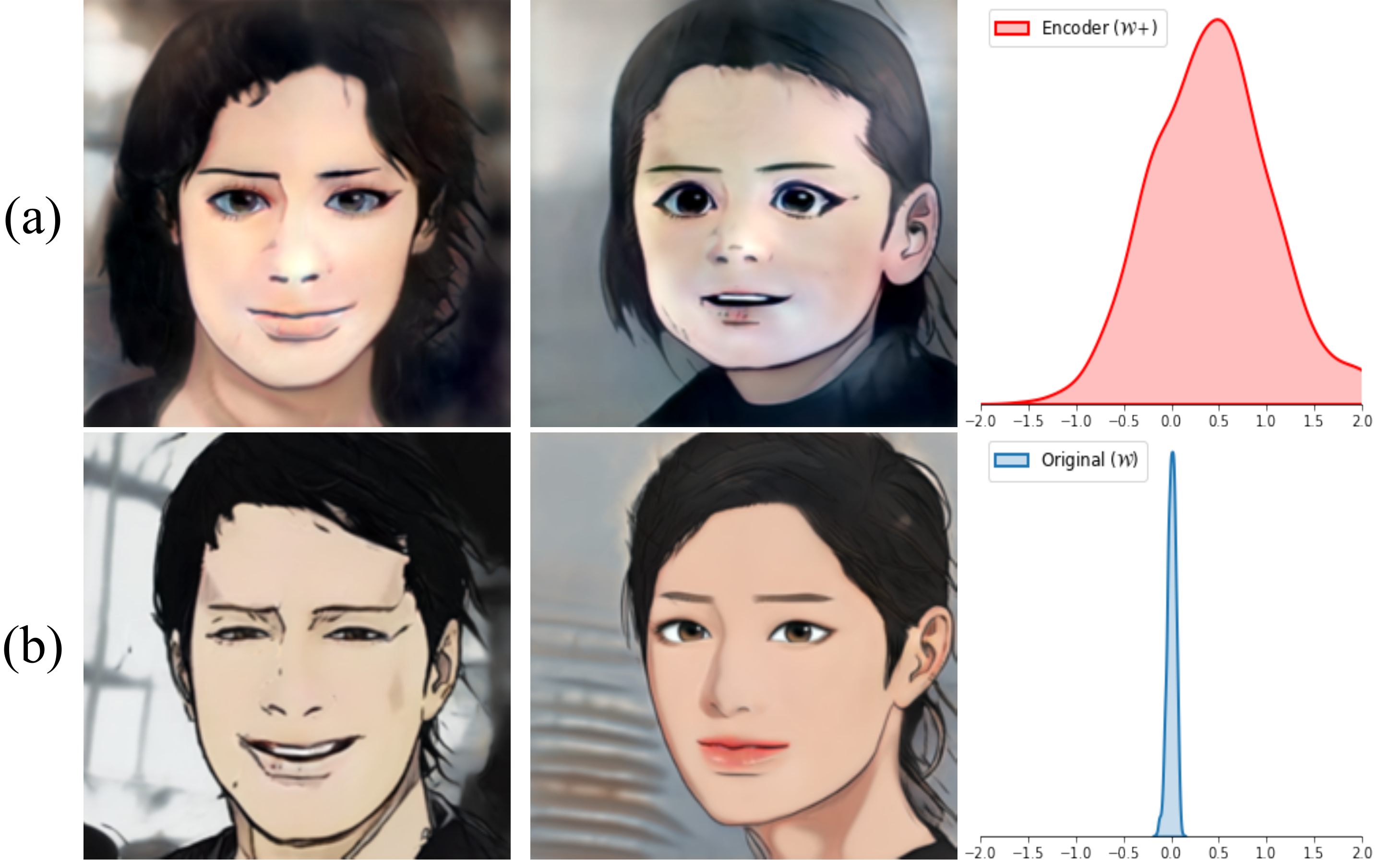}
\caption{Images generated by the same layer-swapped generator from latent codes (a) produced by inversion using the FFHQ pretrained ReStyle encoder, and (b) sampled from the original $\mathcal{W}$ space respectively. The density curves of ${s_{tRGB}}$ for the 4$\times$4 layer which governs the colors of the early resolution~\cite{stylespace} are vastly different (discussed in \Sref{subsec:trgb_replacement}).}

\label{#1}
\end{figure}
}

\newcommand{\figQual}[1]{
\begin{figure*}[t]
\centering
\newcommand{\h}{24mm}
\newcommand{\hsrc}{1.4mm}
\newcommand{\himg}{-0.6mm}
\makebox[\h][c]{\textbf{Source}}\hspace{\himg}
\makebox[\h][c]{\textbf{Target character}}\hspace{\hsrc}
\makebox[\h][c]{(a) CycleGAN}\hspace{\himg}
\makebox[\h][c]{(b) U-GAT-IT}\hspace{\himg}
\makebox[\h][c]{(c) Toonify}\hspace{\himg}
\makebox[\h][c]{(d) AgileGAN}\hspace{\himg}
\makebox[\h][c]{\textbf{(e) Ours}}\hspace{\himg}
\\
\includegraphics[width=\h]{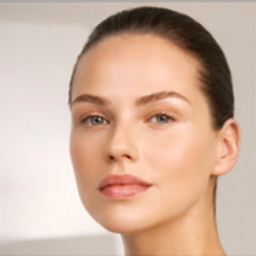}\hspace{\himg}
\includegraphics[width=\h]{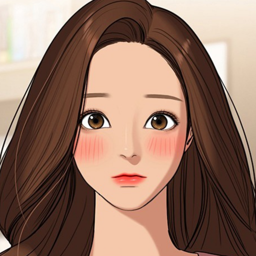}\hspace{\hsrc}
\includegraphics[width=\h]{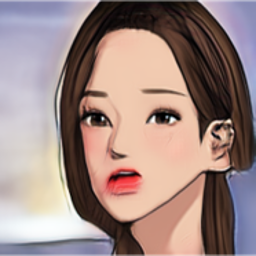}\hspace{\himg}
\includegraphics[width=\h]{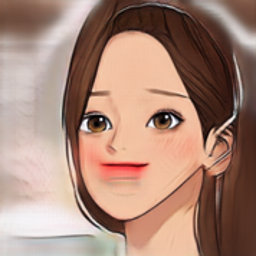}\hspace{\himg}
\includegraphics[width=\h]{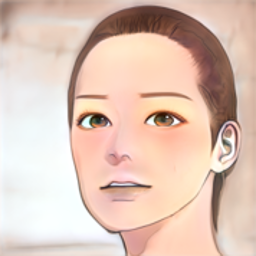}\hspace{\himg}
\includegraphics[width=\h]{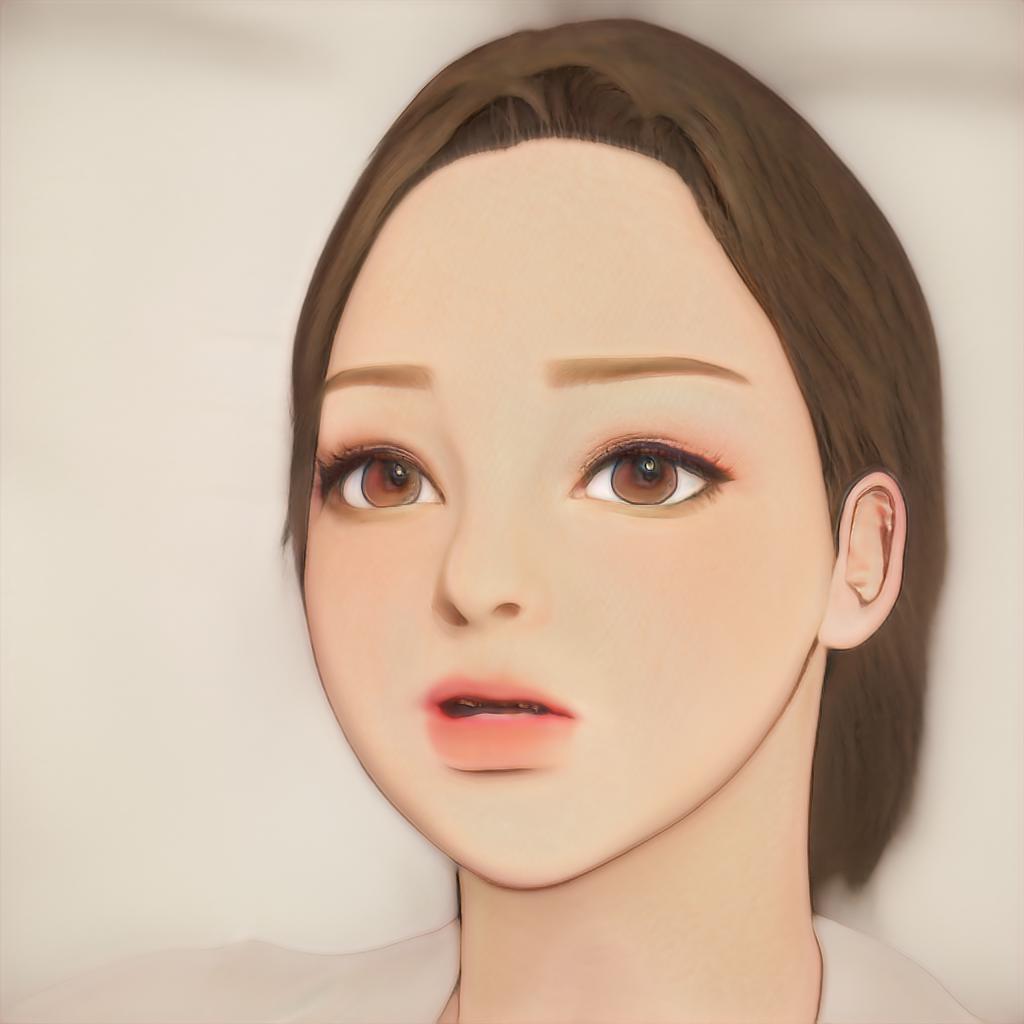}\hspace{\himg}
\includegraphics[width=\h]{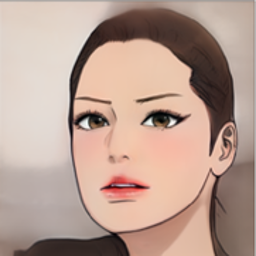}
\\
\includegraphics[width=\h]{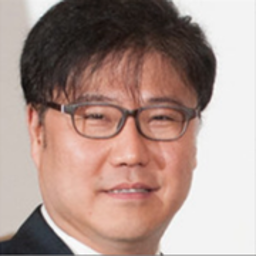}\hspace{\himg}
\includegraphics[width=\h]{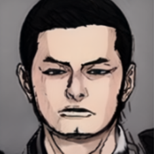}\hspace{\hsrc}
\includegraphics[width=\h]{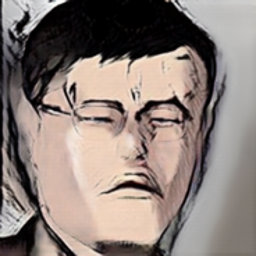}\hspace{\himg}
\includegraphics[width=\h]{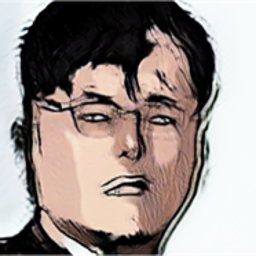}\hspace{\himg}
\includegraphics[width=\h]{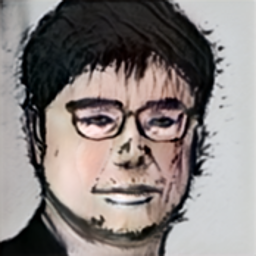}\hspace{\himg}
\includegraphics[width=\h]{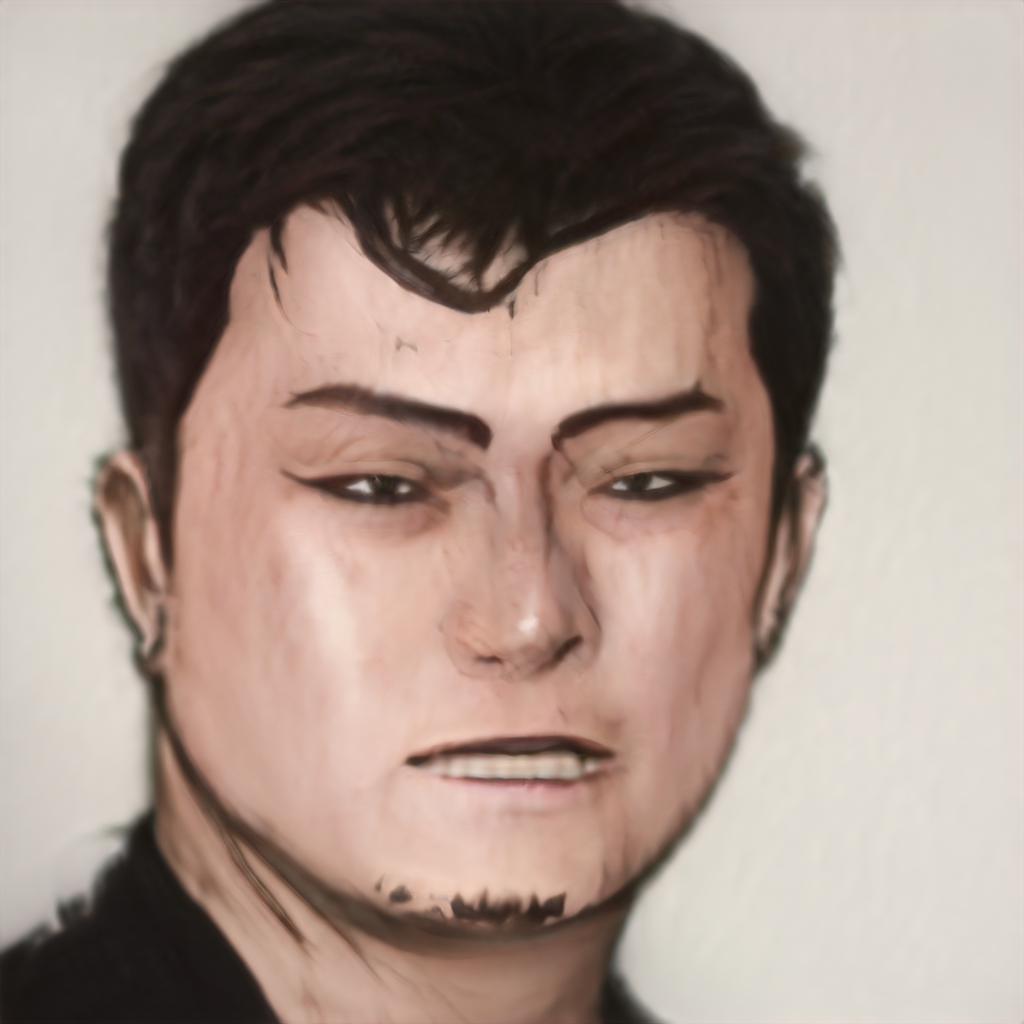}\hspace{\himg}
\includegraphics[width=\h]{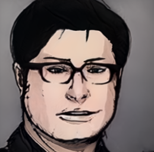}
\\
\includegraphics[width=\h]{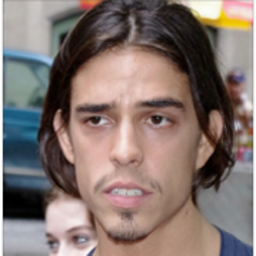}\hspace{\himg}
\includegraphics[width=\h]{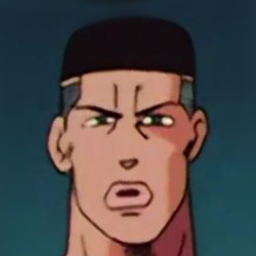}\hspace{\hsrc}
\includegraphics[width=\h]{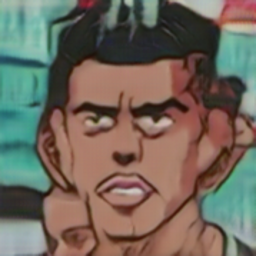}\hspace{\himg}
\includegraphics[width=\h]{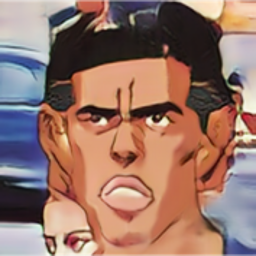}\hspace{\himg}
\includegraphics[width=\h]{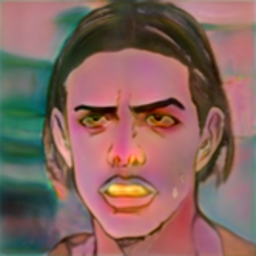}\hspace{\himg}
\includegraphics[width=\h]{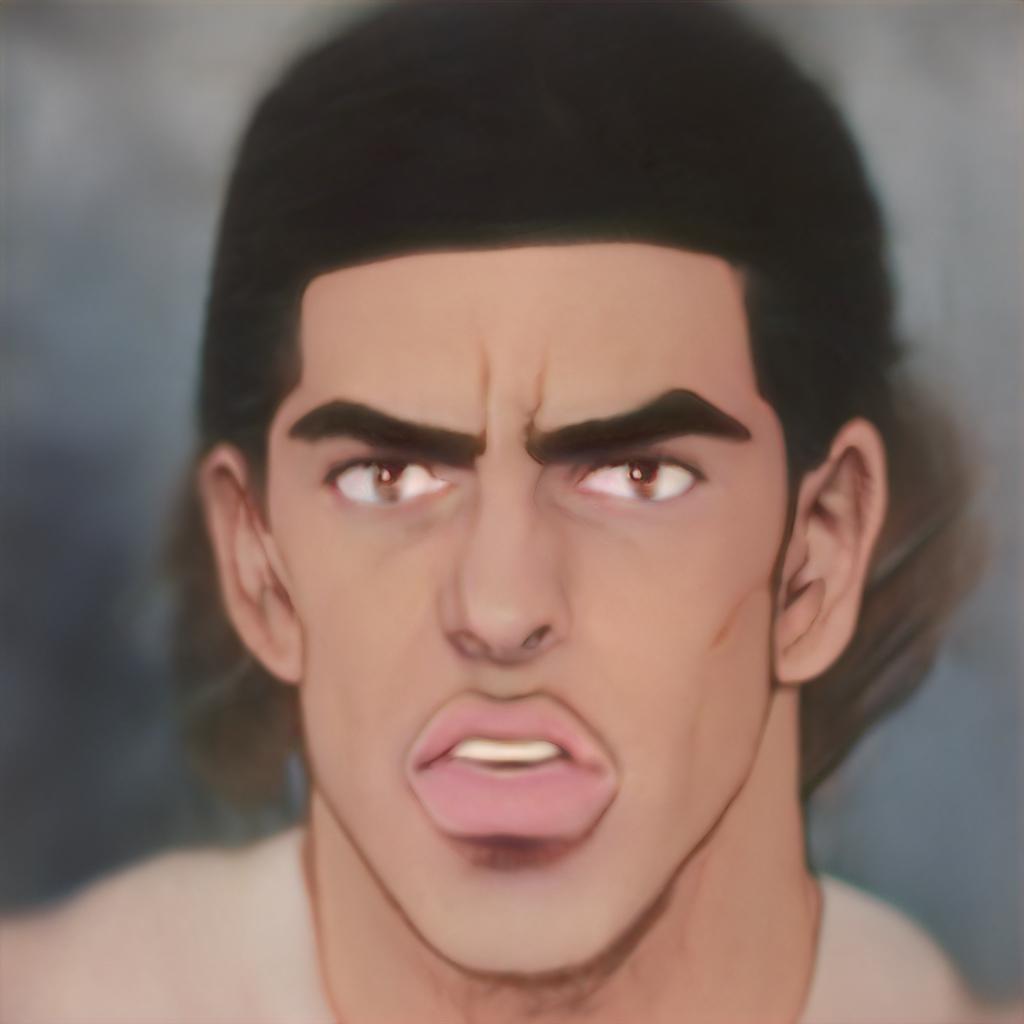}\hspace{\himg}
\includegraphics[width=\h]{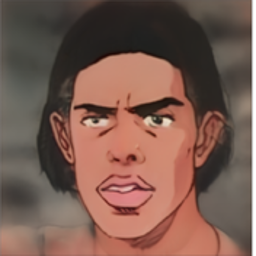}
\caption{Qualitative comparisons. Our model not only effectively maintains the facial structure of the source images but also adequately captures the abstract texture and style from the target character ID. The source images are in the public domain~\cite{whitefemale, asianmaleglass, whitemale}.}
\label{#1}
\end{figure*}
}

\newcommand{\figAblation}[1]{
\begin{figure*}[ht]
\centering
\newcommand{\h}{25mm}
\newcommand{\hsrc}{2mm}
\newcommand{\himg}{-0.4mm}
\makebox[\h][c]{\textbf{Source}}\hspace{\himg}
\makebox[\h][c]{\textbf{Target character}}\hspace{\hsrc}
\makebox[\h][c]{(a)}\hspace{\himg}
\makebox[\h][c]{(b)}\hspace{\himg}
\makebox[\h][c]{(c)}\hspace{\himg}
\makebox[\h][c]{(d)}\hspace{\himg}
\\
\includegraphics[width=\h]{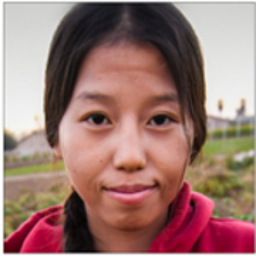}\hspace{\himg}
\includegraphics[width=\h]{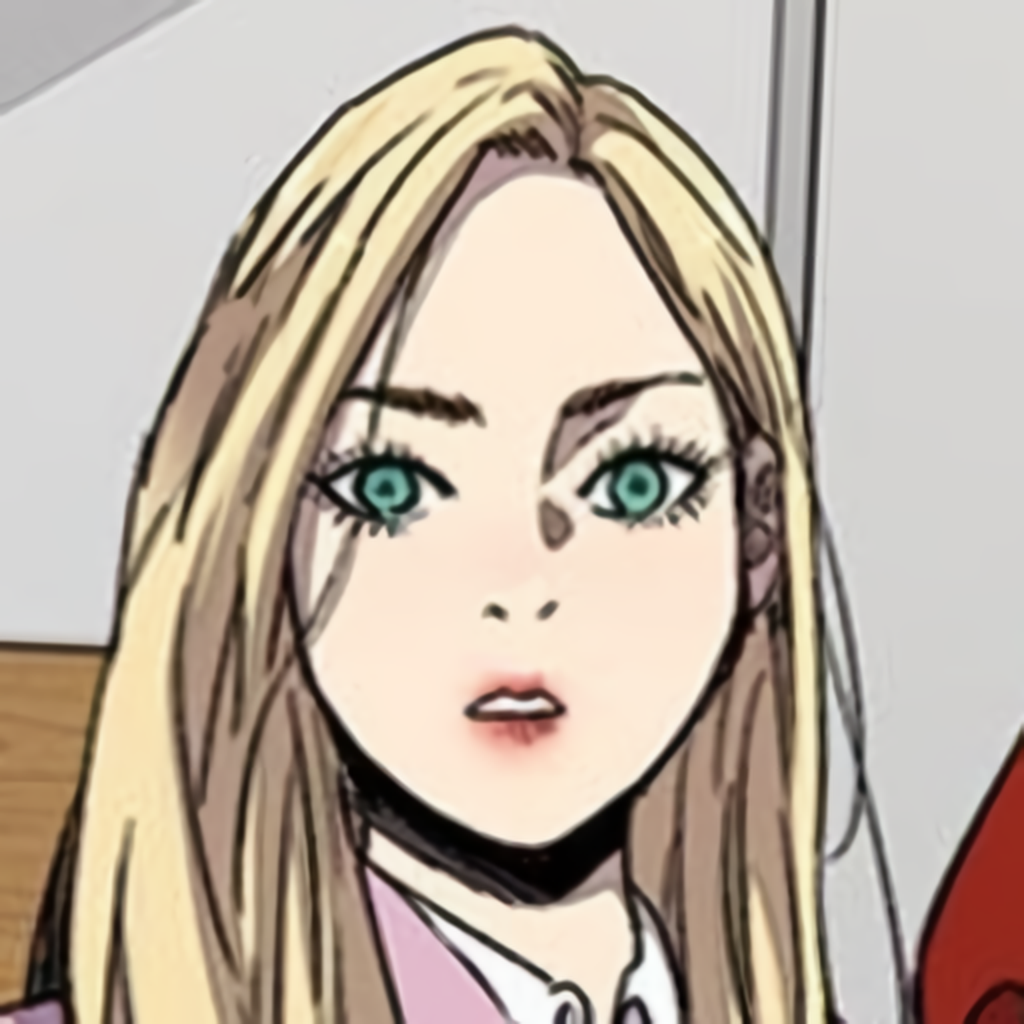}\hspace{\hsrc}
\includegraphics[width=\h]{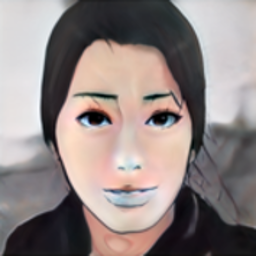}\hspace{\himg}
\includegraphics[width=\h]{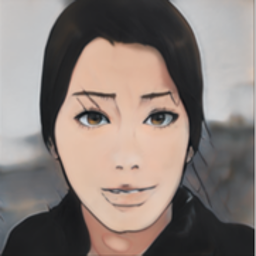}\hspace{\himg}
\includegraphics[width=\h]{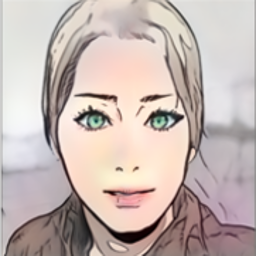}\hspace{\himg}
\includegraphics[width=\h]{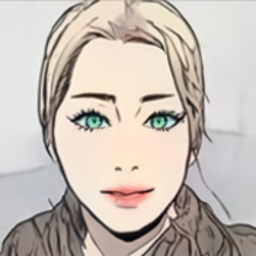}
\\
\includegraphics[width=\h]{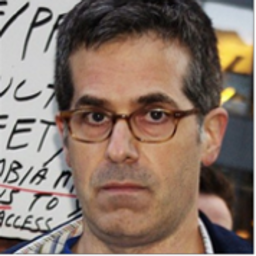}\hspace{\himg}
\includegraphics[width=\h]{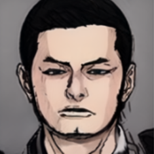}\hspace{\hsrc}
\includegraphics[width=\h]{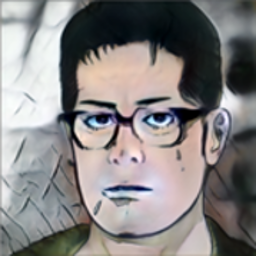}\hspace{\himg}
\includegraphics[width=\h]{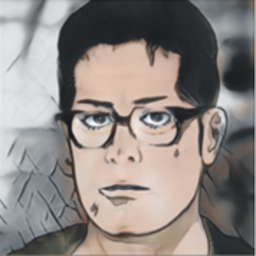}\hspace{\himg}
\includegraphics[width=\h]{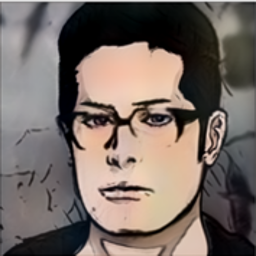}\hspace{\himg}
\includegraphics[width=\h]{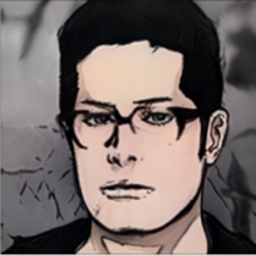}
\\
\includegraphics[width=\h]{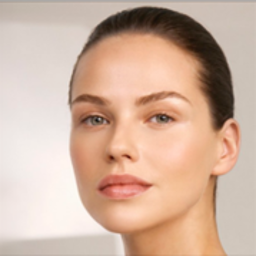}\hspace{\himg}
\includegraphics[width=\h]{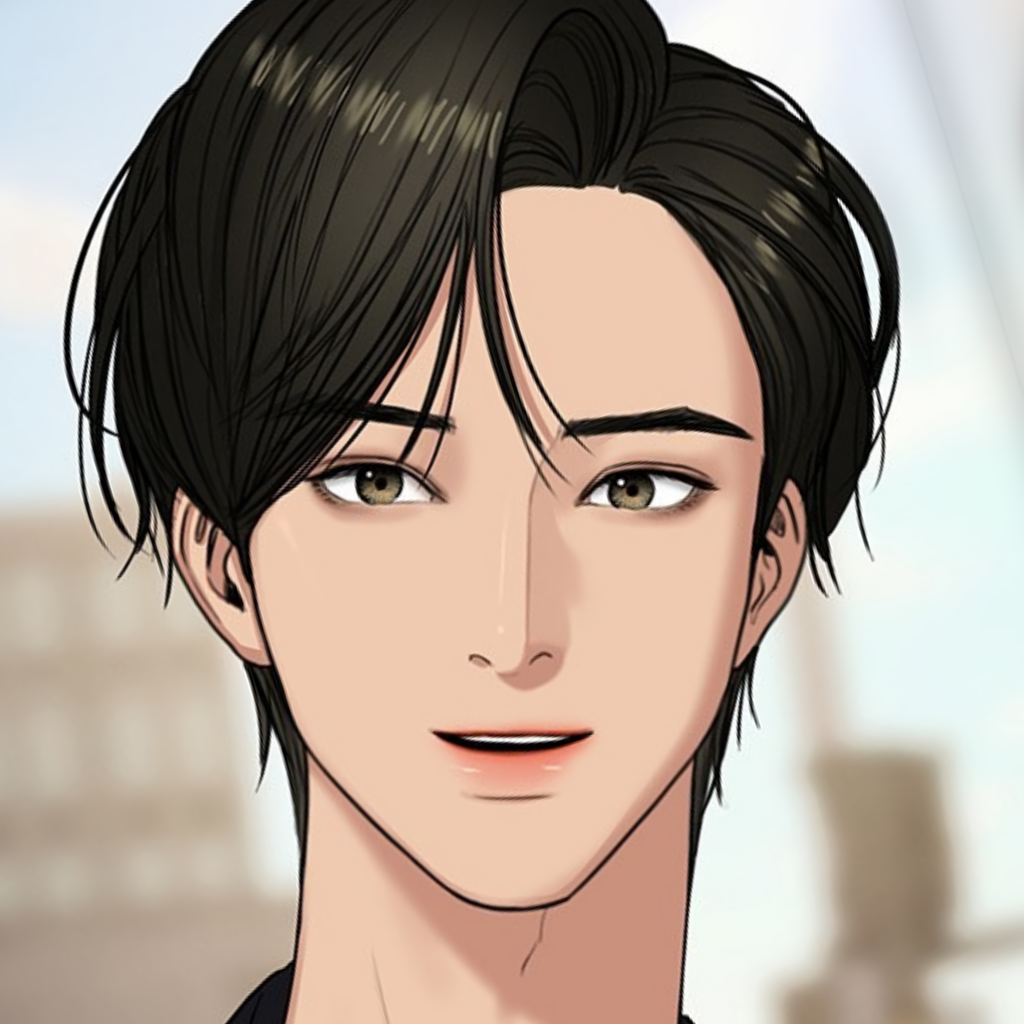}\hspace{\hsrc}
\includegraphics[width=\h]{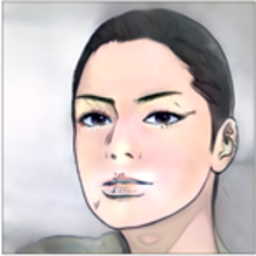}\hspace{\himg}
\includegraphics[width=\h]{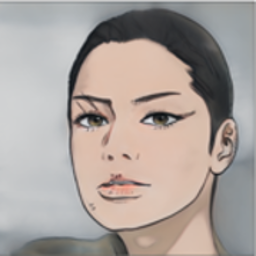}\hspace{\himg}
\includegraphics[width=\h]{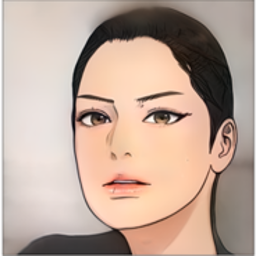}\hspace{\himg}
\includegraphics[width=\h]{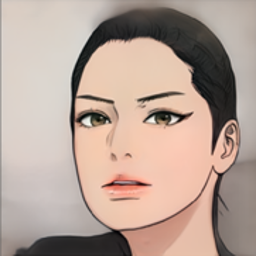}
\caption{Ablation study. (a) Images generated without both style mixing in StyleSpace and tRGB replacement, equivalent to the layer swapping method. (b) Images with only tRGB replacement. (c) Images with only style mixing in the StyleSpace. (d) Full model. Our Cross-domain Style mixing method effectively performs stylization into the desired character style.}
\label{#1}
\end{figure*}
}

\newcommand{\tableQuant}[1]{
\begin{table}[t]
\footnotesize
\caption{Quantitative comparison. {\small\# Nets.} denotes the number of networks required to stylize faces into three characters.}
\begin{center}
\begin{tabular}{cl|c|rrc}
\hline
\multirow{2}{*}{Category} & \multirow{2}{*}{Method} & \multirow{2}{*}{\# Nets.} & \multirow{2}{*}{FID $\downarrow$} & \multirow{2}{*}{KID $\downarrow$} & User \\
& & & & & Score $\uparrow$ \\
\hline\hline
\multirowcell{2}[0pt][r]{I2I-\\based} & CycleGAN & 3 & \textbf{59.77} & \textbf{14.35} & \hspace{0.4em}2.25 \\
& U-GAT-IT & 3 & 59.96 & 15.23 & \hspace{0.4em}4.50 \\\hline
\multirowcell{3}[0pt][r]{StyleGAN-\\based} & Toonify & 3 & 103.45 & 54.71 & \hspace{0.4em}6.50\\
& AgileGAN & 3 & 158.32 & 134.19 & 26.00\\
& \textbf{Ours} & \textbf{1} & \textbf{85.72} & \textbf{40.70} & \textbf{60.75}\\
\hline
\end{tabular}
\end{center}
\label{#1}
\end{table}
}

\newcommand{\figColorDistort}[1]{
\begin{figure}[t]
\centering
\newcommand{\h}{26mm}
\newcommand{\himg}{1.5mm}
\makebox[\h][c]{(a)}\hspace{\himg}
\makebox[\h][c]{(b)}\hspace{\himg}
\makebox[\h][c]{(c)}
\\
\includegraphics[width=\h]{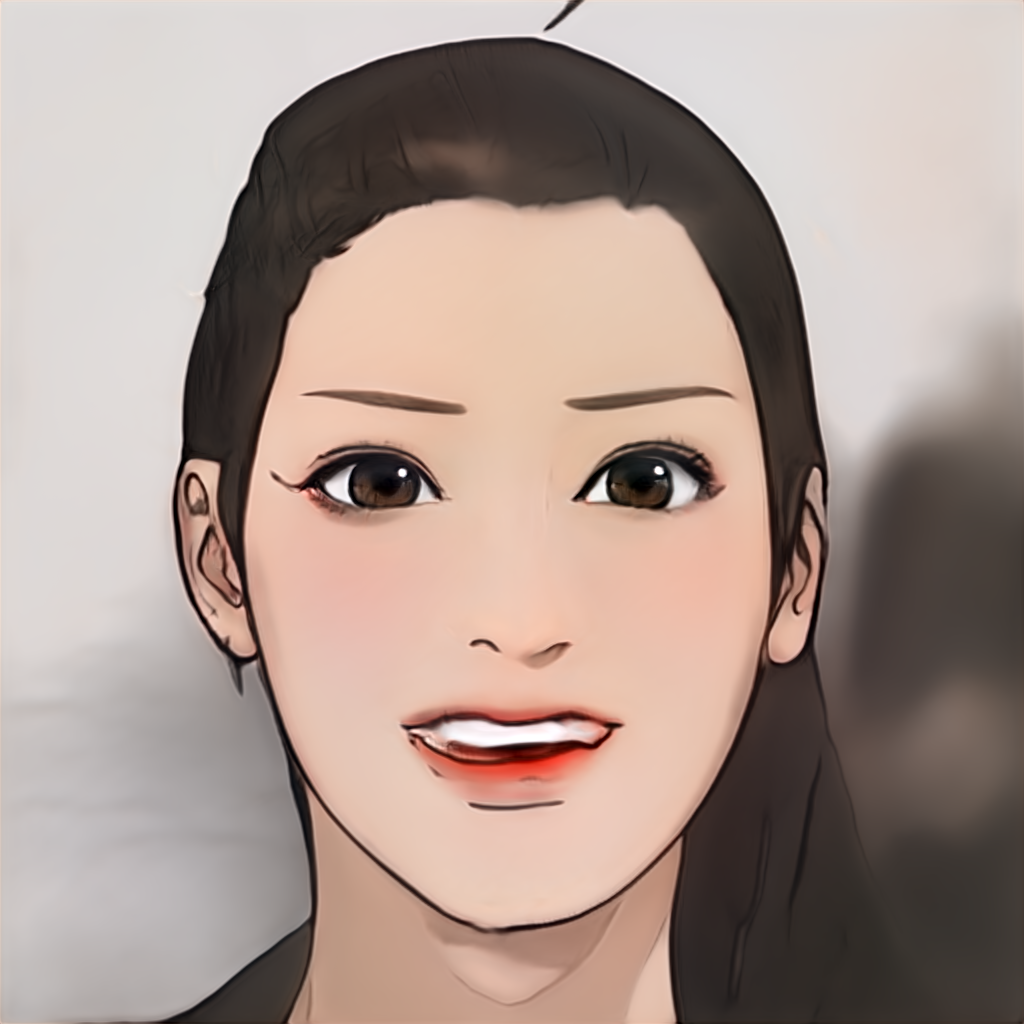}\hspace{\himg}
\includegraphics[width=\h]{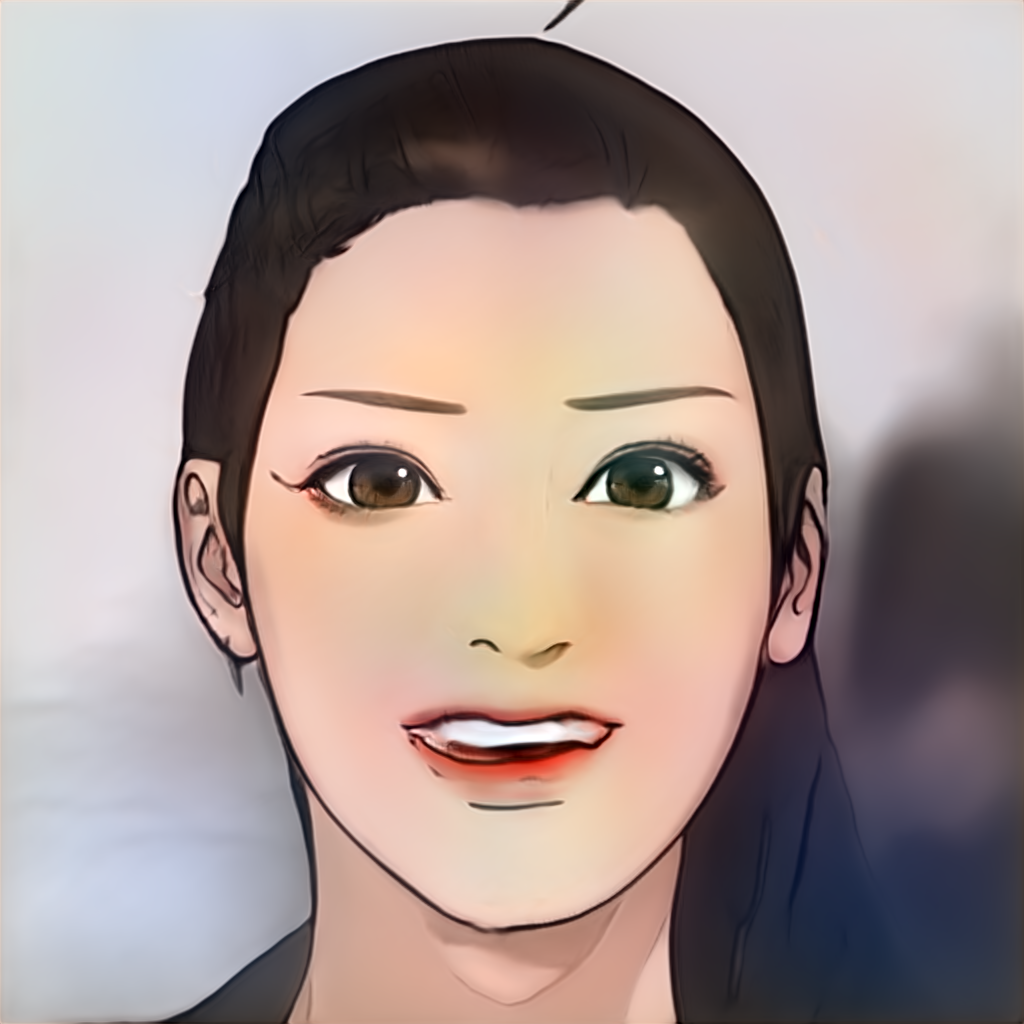}\hspace{\himg}
\includegraphics[width=\h]{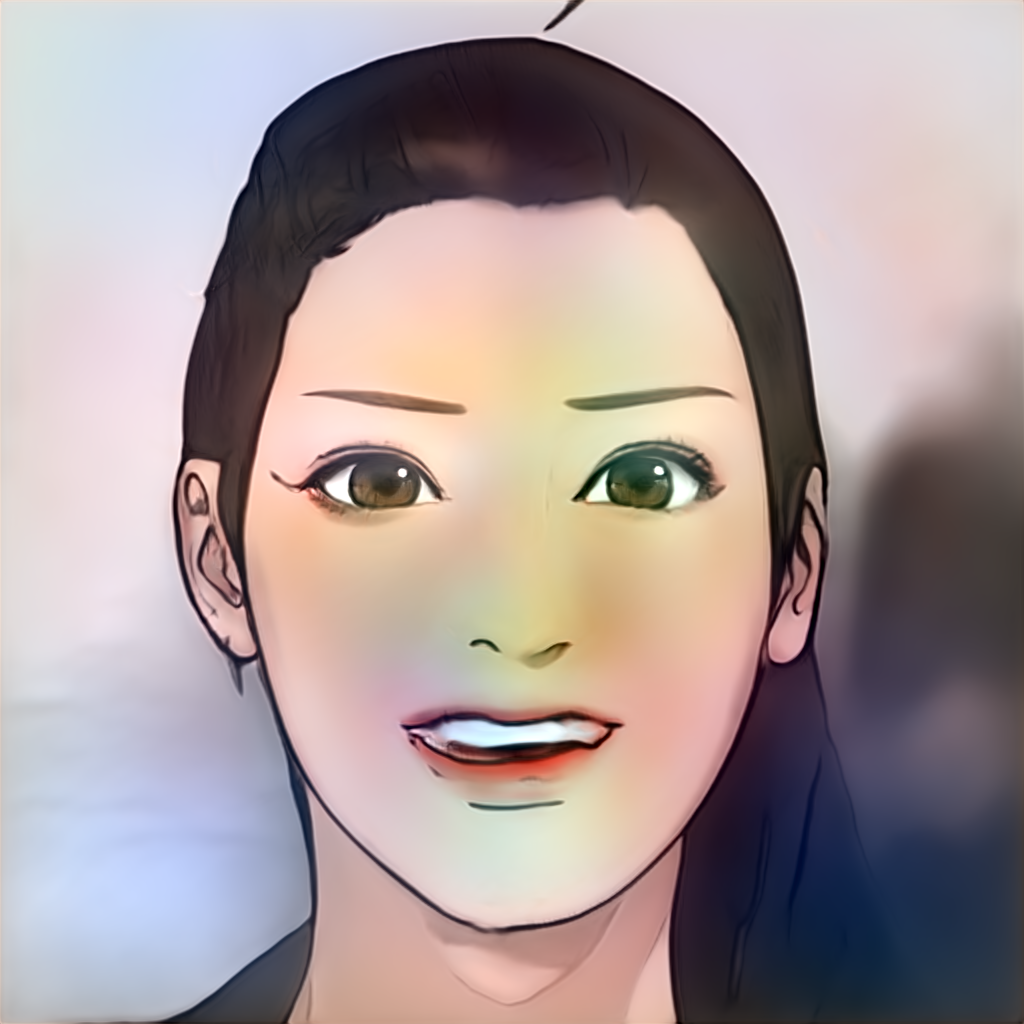}
\caption{Color distortion experiment. (a) Result with tRGB replacement (no $s_{tRGB}$ perturbation). (b, c) Results with $s_{tRGB}$ perturbation ($N=10, 20$). It show that the altered tRGB parts only affect the color distribution without changing facial features.}
\label{fig:color_distort}
\end{figure}
}

\newcommand{\figCompLatentspace}[1]{
\begin{figure}[t]
\centering
\newcommand{\h}{26mm}
\newcommand{\himg}{1.5mm}
\makebox[\h][c]{(a)}\hspace{\himg}
\makebox[\h][c]{(b)}\hspace{\himg}
\makebox[\h][c]{(c)}
\\
\includegraphics[width=\h]{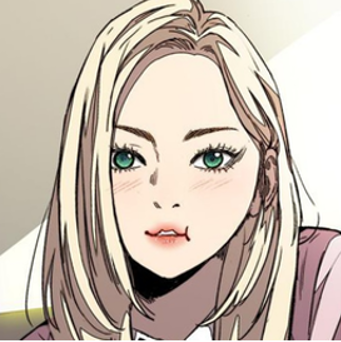}\hspace{\himg}
\includegraphics[width=\h]{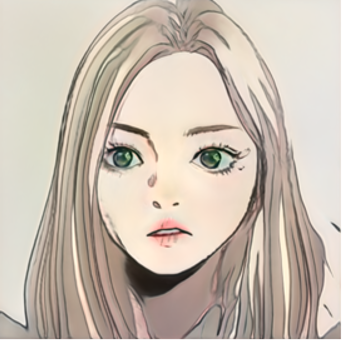}\hspace{\himg}
\includegraphics[width=\h]{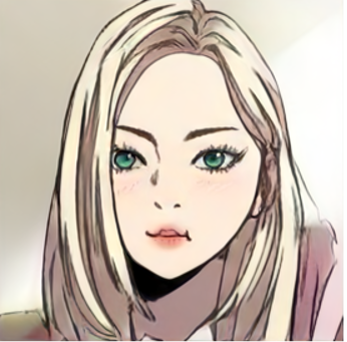}
\caption{Latent space analysis. (a) Input cartoon image. (b, c) Images generated from the inverted latent code in the (b) $\mathcal{W}$ and (c) $\mathcal{W+}$ spaces. The $\mathcal{W+}$ space (c) clearly shows better reconstruction quality while adequately preserving the cartoon texture.}
\label{fig:comp_latentspace}
\end{figure}
}

\newcommand{\figFailure}[1]{
\begin{figure}[t]
\centering
\newcommand{\h}{20mm}
\newcommand{\himg}{-0.6mm}
\newcommand{\hsrc}{2.0mm}
\includegraphics[width=\h]{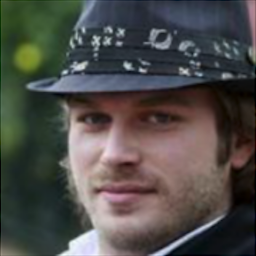}\hspace{\himg}
\includegraphics[width=\h]{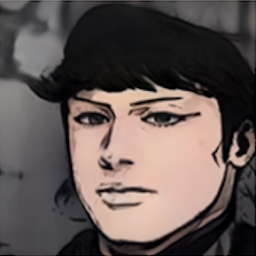}\hspace{\hsrc}
\includegraphics[width=\h]{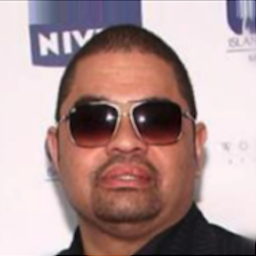}\hspace{\himg}
\includegraphics[width=\h]{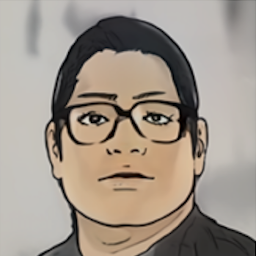}\hfill
\caption{Failure cases.}
\label{#1}
\end{figure}
}

\newcommand{\figApplication}[1]{
\begin{figure}[t]
\newcommand{\h}{19.5mm}
\newcommand{\hOur}{40mm}
\newcommand{\hsrc}{1.0mm}
\newcommand{\himg}{-0.6mm}
\makebox[\h][c]{\textbf{Source}}\hspace{\hsrc}
\makebox[\h][c]{(a)}\hspace{\hsrc}
\makebox[\hOur][c]{(b)}\hspace{\himg}
\\
\includegraphics[width=\h]{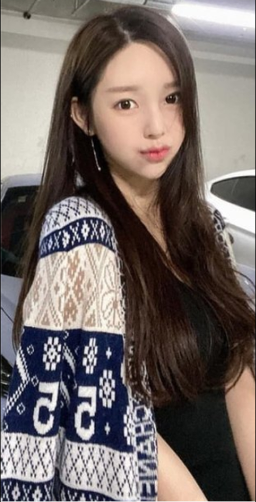}\hspace{\hsrc}
\includegraphics[width=\h]{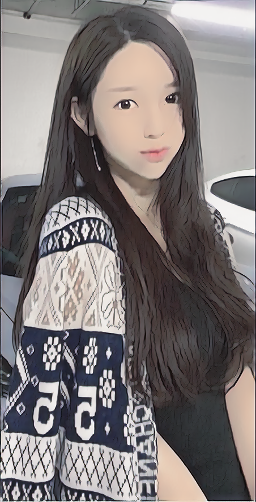}\hspace{\hsrc}
\includegraphics[width=\h]{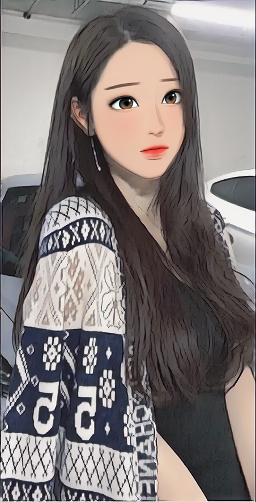}\hspace{\himg}
\includegraphics[width=\h]{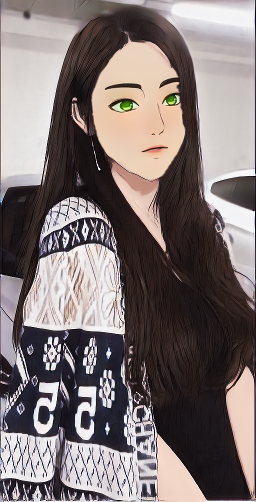}
\caption{The photo-to-cartoon application. (a) Result of the previous cartoonization method trained on both the character face and scene datasets. (b) Results of our proposed approach.}
\label{fig:application}
\end{figure}
}

\newcommand{\figColorDensity}[1]{ 
\begin{figure}[t]
\newcommand{\h}{50mm}
\newcommand{\himg}{5mm}
\centering
\includegraphics[width=\h]{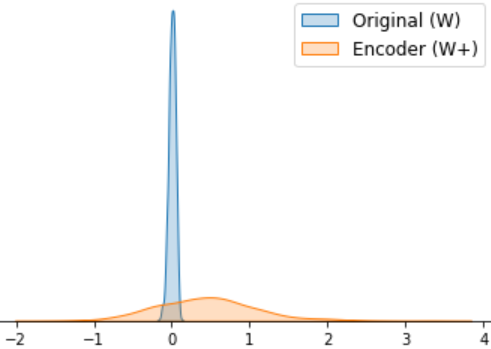}
\caption{Density curves of $s_{tRGB}$ (\textit{e.g.,} the channel index 1) from two sets of latent codes. Both sets are averaged by each latent code. The encoder shows a vastly different $s_{tRGB}$ density, leading to a different color distribution in the output image.}
\label{fig:color_density}
\end{figure}
}

\newcommand{\figConsistency}[1]{
\begin{figure}[t]
\newcommand{\h}{83mm}
\includegraphics[width=\h]{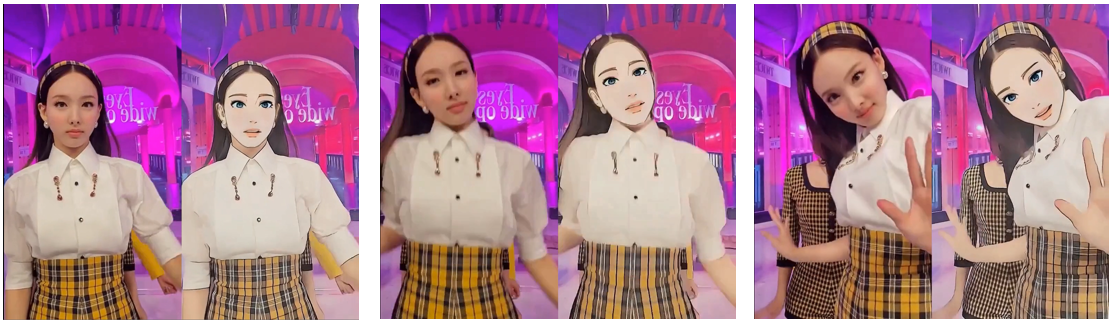}
\caption{The video-to-cartoon application. Our method shows consistent character styles over
video frames, owing to the explicit style injection through style mixing.}
\label{fig:consistency}
\end{figure}
}

\newcommand{\figtRGB}[1]{
\begin{figure}[t]
\centering
\includegraphics[width=0.5\linewidth]{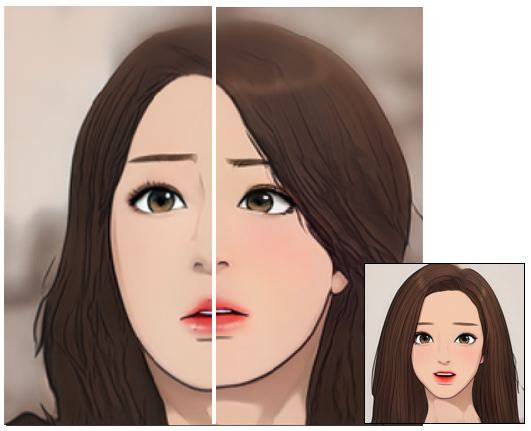}
\caption{Stylized results with ${s^w_{tRGB}}$ from $s^{w}$ with a randomly sampled $w$ in the original latent space (Left), and with ${s^c_{tRGB}}$ from $s^{c}$ with a specific target character $c$ (Right). It is shown that ${s^c_{tRGB}}$ helps stylize the unique makeup-look in the character.}
\label{fig:figtRGB}
\end{figure}
}

\newcommand{\figVideo}[1]{
\begin{figure*}[t]
\centering
\includegraphics[width=\linewidth]{figure/suppl/video-to-cartoon/result.png}
\caption{Video-to-cartoon application.}
\label{fig:figVideo}
\end{figure*}
}

\newcommand{\figResults}[1]{
\begin{figure*}[h]
\centering
\includegraphics[width=\linewidth]{figure/suppl/results/results.png}
\caption{Stylized results from a single generator.}
\label{fig:figResults}
\end{figure*}
}

\newcommand{\figPhoto}[1]{
\begin{figure*}[h]
\centering
\includegraphics[width=\linewidth]{figure/suppl/photo-to-cartoon/results.png}
\caption{Photo-to-cartoon application.}
\label{fig:figPhoto}
\end{figure*}
}

\begin{document}
\title{Cross-Domain Style Mixing for Face Cartoonization}

\author{
Seungkwon Kim \qquad Chaeheon Gwak \qquad Dohyun Kim \\ Kwangho Lee \qquad Jihye Back \qquad Namhyuk Ahn* \qquad Daesik Kim* \\ NAVER WEBTOON AI
}

% teaser image
\twocolumn[{
\renewcommand\twocolumn[1][]{#1}
\maketitle
\begin{center}
\centering
\captionsetup{type=figure}
\includegraphics[width=\linewidth]{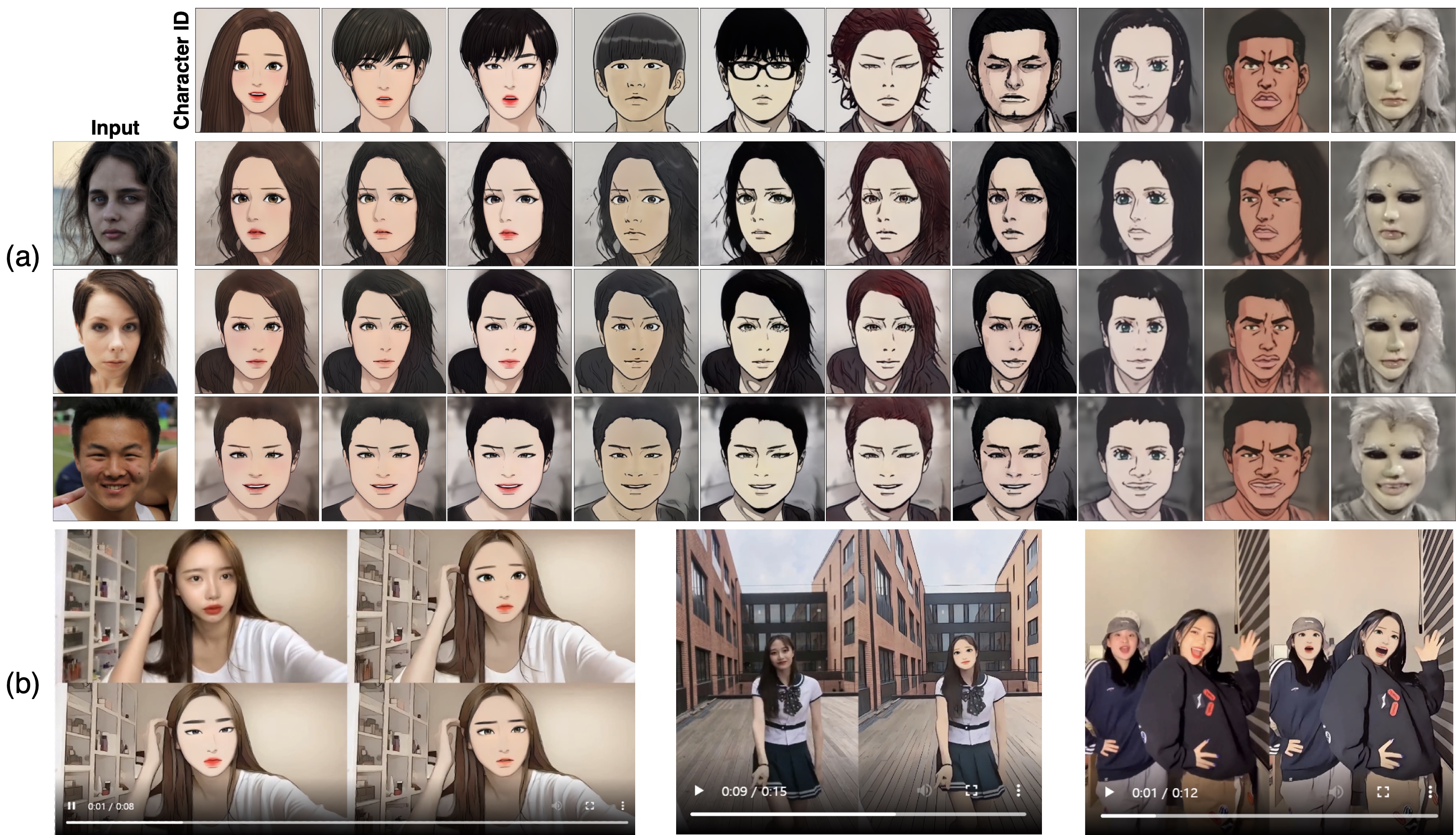}
\captionof{figure}{(a) Our method can stylize faces into multiple cartoon characters at different abstraction levels with a single generator by simply switching the character IDs. (b) Our proposed method can also be extended to a video-to-cartoon application with a few extra modules.}
\label{fig:teaser}
% The input images are in the public domain~\cite{img5, img2, img1}
\end{center}
}]

% \twocolumn[{
% \renewcommand\twocolumn[1][]{#1}
% \maketitle
% \begin{center}
% \centering
% \captionsetup{type=figure}
% \includegraphics[width=\linewidth]{figure/teaser.png}
% \captionof{figure}{(a) Our method can stylize faces into multiple cartoon characters at different abstraction levels using only a single generator by simply switching the character IDs. (b) Our proposed method can also be extended to a video-to-cartoon application with a few extra modules. The input images are in the public domain~\cite{img5, img2, img1}.}
% \label{fig:teaser}
% \end{center}
% }]

\begin{abstract}
Cartoon domain has recently gained increasing popularity.
%owing to its accessibility and availability on digital platforms.
Previous studies have attempted quality portrait stylization into the cartoon domain; however, this poses a great challenge since they have not properly addressed the critical constraints, such as requiring a large number of training images or the lack of support for abstract cartoon faces.
Recently, a layer swapping method has been used for stylization requiring only a limited number of training images; however, its use cases are still narrow as it inherits the remaining issues. In this paper, we propose a novel method called Cross-domain Style mixing, which combines two latent codes from two different domains. Our method effectively stylizes faces into multiple cartoon characters at various face abstraction levels using only a single generator without even using a large number of training images.
\end{abstract}
\customfootnotetext{*}{~indicates equal contribution.}

\section{Introduction}
\label{sec:introduction}

% \figOverviewFramework{fig:overview_framework}

Cartoons, which are a composite art form that combines painting and storytelling, have a tremendous number of readers worldwide. An application can best reveal a reader's immersion in cartoons and his/her empathy for a character by enabling him/her  to become part of the character itself in an external form (\Fref{fig:teaser}). In this respect, portrait stylization applied to the cartoon domain has an enormous demand. The advancement of deep neural networks, particularly generative adversarial networks (GANs)~\cite{gan}, has shown great progress in portrait stylization of the cartoon domain. Among them, image-to-image (I2I) translation is a notable approach~\cite{ugatit,cartoongan}; however, it requires a large number of cartoon images and exhaustive GPU resources. In addition, it often lacks the ability to express character-specific cartoon features, thus limiting its practical application.

To mitigate these issues, Toonify~\cite{toonify} introduced a layer swapping approach that interpolates two StyleGAN2~\cite{stylegan2} generators trained in different domains. It can perform a decent portrait stylization using only hundreds of training images. However, this approach suffers from critical quality issues that prevent it from being a robust application. In the results of Toonify, the cartoon styles cannot be expressed in fine detail and the colors may be distorted in some parts of the output images. These problems deteriorate when the texture and abstraction level of the target cartoon character are vastly different from those of human faces, as commonly seen in the style of Japanese animation.

In this paper, we propose a novel method called Cross-domain Style mixing, which combines latent codes from two different domains. We obtain the latent codes for both the input natural face (source domain) and the cartoon (target domain) images and perform style mixing in the same latent space of the layer-swapped generator (\Fref{fig:overview_framework}).
In detail, we first carefully design inversion strategies for generating the latent codes for the source and target domains. We employ a pretrained encoder (\textit{e.g.,} ReStyle~\cite{restyle}) for the source domain and a projection protocol for the target domain, both in the expressive $\mathcal{W+}$ space~\cite{image2stylegan}. This makes the latent codes suitable for combining. Then, we perform style mixing on them in the $\mathcal{S}$ space (StyleSpace~\cite{stylespace}).

Although it successfully preserves the detailed cartoon features, we observed that there is still color distortion in the output, just like in the previous layer swapping method. Our inspection revealed that the layer-swapped generator tends to produce images with color artifacts when taking latent codes as inputs from the pretrained encoder owing to the domain disparity. Hence, we apply tRGB replacement method, which manipulates parts of the style parameters $s \in \mathcal{S}$ of the input image, enabling the output image to follow the color distribution of the target cartoon domain, and, consequently, it successfully removes the color artifacts. 

To the best of our knowledge, this is the first approach that extends conventional style mixing, which only accepts latent codes from the same domain, to enable it  to deal with cross-domains.
Our method is simple but effective in terms of both quality and stability (\Fref{fig:teaser}). In addition, it enables a single generator to perform one-for-all stylizations for cartoon faces at various abstraction levels. This is because our method utilizes character-wise ``latent codes,'' unlike the previous methods, which need to prepare character-wise ``generators.''
We would also like to note that our framework requires neither additional training tricks (\textit{e.g.,} auxiliary loss or regularization) nor extra architectural adjustments other than fine-tuning the StyleGAN2 generator. Therefore, we believe that it is highly applicable to real-world scenarios owing to its high efficiency in training and deployment.
Our contributions are summarized as follows:
\begin{itemize}
    \item We thoroughly investigated the limitations of the layer swapping method as a cartoon stylization that has various abstract faces and styles. 

    \item We propose a simple but quality stylization framework specialized for the cartoon domain. A single generator can stylize faces into multiple cartoon characters by simply switching the character IDs. It only requires minimal training efforts and a training dataset ($<$100 images for each cartoon character). 
    
    \item We demonstrate that, when our method is combined with a few extra modules, it can even provide superior photo or video cartoonization results.
\end{itemize}
\section{Background}
\label{sec:related work}

\noindent\textbf{Portrait stylization.}
The neural style transfer~\cite{gaty2016}-based approach utilizes a pretrained network to perform image stylization~\cite{selim2016,kaur2019}.
However, it is mostly limited to the geometric deformation of facial features, such as exaggeration in caricatures and comics.
Moreover, the GAN-based image-to-image (I2I) paradigm has also been actively studied in this field~\cite{pix2pix,ugatit}.
Although they can generate attractive stylization results, they only express a limited range of target or input features and rely on a large number of training images, which is not suitable for many real-world scenarios.

Recently, Toonify~\cite{toonify} proposed a layer swapping method to address the aforementioned issues of I2I methods. It swaps the layers of pretrained and fine-tuned StyleGAN2 generators trained in the source and target domains, respectively. The framework shows plausible stylization results using a small number of training images; however, for highly abstract cartoon faces, it often lacks the expression of fine cartoon textures and yields undesired color artifacts, thus limiting its use cases to specific cartoon characters only.
In contrast, our method successfully produces fine textures and features of diverse cartoon characters while resolving the unpleasant color artifacts.
\smallskip

\noindent\textbf{GAN inversion.} The inversion of an image into the latent space of a GAN generator has been an active research area~\cite{ganinvsurvey}. In general, there are two types of approaches for GAN inversion. One is the optimized-based, which utilizes optimization algorithms to update the latent code such that the loss between the synthesized and the target image is minimized~\cite{stylegan, image2stylegan}. The  other is the encoder-based approach, which uses a network to map an image into the latent space in a deterministic manner~\cite{psp, e4e, restyle}. In our work, we use the ReStyle encoder~\cite{restyle} for the source domain because of its superior reconstruction and expression abilities by employing the $\mathcal{W+}$ space~\cite{image2stylegan}.

\section{Motivation \& Pre-analysis}  
\label{subsec:motivation}

Although the layer swapping approach produces decent stylization results, we observed that there are two major quality issues when it is applied to the cartoon domain.

First, the layer-swapped generator cannot generate images with faithful cartoon features. When a StyleGAN2 generator produces images sampled from the original $\mathcal{W}$ space, it has the highest fidelity because the space represents the training distribution~\cite{stylegan,stylegan2}. On the other hand, for a layer-swapped generator, the original $\mathcal{W}$ space is aligned only with the low-resolution layers as these layers and the mapping network come from the FFHQ fine-tuned generator; however, the same $\mathcal{W}$ space becomes unaligned with the high-resolution layers trained in the cartoon domain. This results in output images lacking detailed cartoon features to be synthesized from the high-resolution layers, thus losing the character-wise identities (\Fref{fig:comp_g}c).

% \figCompColor{fig:comp_color}
Second, color distortion exists in the output images. The layer swapping method uses a pretrained encoder to invert a given natural face image into the latent code. We observed that this can cause an unexpected quality degradation, only confined to colors, because the encoder pretrained on the FFHQ dataset only represents color ranges in the source domain, not in the target cartoon domain (\Fref{fig:comp_color}a).

To mitigate these issues, previous studies adopted latent-consistent fine-tuning and VAE encoder~\cite{agilegan} or specifically trained encoder for the layer-swapped generator~\cite{psp}.
However, these approaches assume that the cartoon and natural faces share many features and textures; therefore, they use similarity loss (\textit{e.g.,} perceptual loss between the natural and cartoon faces) as a strong regularization. We observed that such regularization results in outputs that do not apply to the highly abstract or flat cartoon faces, excessively preserving textures and features of the natural faces (\Fref{fig:qual}d).

\figCompGenerator{fig:comp_g}

\section{Method}
\label{sec:method}

We propose a novel portrait stylization method, namely, the Cross-domain Style mixing method, that not only overcomes the limitations of I2I methods but also tackles two critical issues of the layer swapping method, as discussed in \Sref{subsec:motivation}.
Unlike those in previous methods~\cite{toonify, agilegan,psp}, our model can express a diverse range of cartoon styles using only a single generator, which provides superior practicality in real-world scenarios.

\figOverviewFramework{fig:overview_framework}

Previous style mixing methods take two latent codes from the same domain and decode the output through a pretrained generator.
In contrast, we combine the latent codes that belong to different domains (\textit{e.g.,} natural and cartoon faces) and decodes the result using a layer-swapped generator. To do that, we use different inversion methods for each domain and perform style mixing in the $\mathcal{S}$ space (StyleSpace~\cite{stylespace}). We also apply the \texttt{tRGB} replacement method to address the color distortion issue (\Fref{fig:overview_framework}).

\subsection{Framework}
\label{subsec:framework}

Our method is simple; removing all redundant modeling tricks and architectural complexities introduced in previous studies; accordingly, portrait stylization can be easily performed with minimal effort and resources.

Let the natural face image be $x^{f}$ and the cartoon image of a target character ID $c$ be $x^{c}$. Our goal is to generate $\hat{x}^c$ by stylizing the input $x^f$ into the desired character $c$.
We first construct a layer-swapped generator $G_{swap}$ as the layer swapping method, replacing the high-resolution layers of the FFHQ pretrained generator with those of the fine-tuned generator trained on the cartoon dataset. 

Next, we prepare the character ID-specific latent codes $w^c$ by randomly sampling $k$ cartoon images $x_{1,2,...k}^c$ for each character ID $c$ from the cartoon dataset and inverting them into the latent space of $G_{swap}$ using an inversion method for the target domain ($Inv^{tgt}(.)$).
\begin{equation}
\label{eq:eq1}
w^{c} = \frac{1}{k}\sum_{i=1}^k Inv^{tgt}\left(x_{i}^{c}\right)
\end{equation}

For the inference step, the input face image $x^{f}$ is first inverted into the latent code $w^{f}$ by the source domain-specific encoder trained on the FFHQ dataset, denoted as $Inv^{src}(.)$.
\begin{equation}
\label{eq:2}
w^{f} = Inv^{src}(x^f)
\end{equation}
Note that the encoder generates $w^{f}$ in the same latent space as $w^{c}$ because $w^{c}$ is inverted using $G_{swap}$, which adopts the mapping network from the FFHQ pretrained generator. Both $w^{f}$ and $w^{c}$ are consequently eligible for style mixing despite the domain disparity. Finally, we generate the output $\hat{x}^{c}$ using our proposed Cross-domain Style mixing method ($CDSM(.)$) and $G_{swap}$ as follows:
\begin{equation}
\label{eq:3}
\hat{x}^{c} = G_{swap}\left(CDSM(w^{f}, w^{c}, m)\right).
\end{equation}
Here, $m$ is the style mixing level that determines the strength of the cartoon style in the output image. 
$CDSM(.)$ is composed of the style mixing in StyleSpace method $SM(.)$ and the tRGB replacement method $R(.)$ and is defined as follows:
\begin{equation}
\begin{split}
CDSM(w^f, w^c, m) &\equiv SM\left(R\left(s^f, s^c\right), s^c, m\right), \\
s^{f} = A(w^{f}), &\quad s^{c} = A(w^{c})
\label{eq:cdsm}
\end{split}
\end{equation}
where $A$ indicates the layer-wise affine layers in the StyleGAN2 generator, which provides less entangled style parameters $s \in \mathcal{S}$, as suggested in StyleSpace.

% \figQual{fig:qual}
% \figCompColor{fig:comp_color}

\subsection{Style mixing in StyleSpace}
\label{subsec:stylemixing}
We first describe the style mixing in StyleSpace method $SM(.)$ in our framework. Unlike the conventional style mixing methods, our method utilizes two different inversion techniques and then performs style mixing in the $\mathcal{S}$ space.
\smallskip

\noindent\textbf{Inversion for the natural image.}
In particular, we choose the Restyle encoder~\cite{restyle} in our framework to obtain $w^{f}$ because the encoder is based on the $\mathcal{W+}$ space, which is known for its superior reconstruction and feature preservation~\cite{image2stylegan}. Furthermore, its feed-forward inference nature makes it ideal for many practical applications. However, we found that, with this encoder, the layer-swapped generator unexpectedly produces color-distorted images. This issue is fully addressed by the tRGB replacement method (\Sref{subsec:trgb_replacement}) in our framework, which ensures the robust use of the encoder without quality deterioration.
\smallskip

\noindent\textbf{Inversion for the cartoon image.}
Generating latent codes for cartoon images $w^{c}$ requires a different inversion scheme. First, the codes can be processed in advance for inference as they are all stored in a look-up table and later queried by a request with the desired character ID. Second, training an encoder is not trivial because only a limited number of training images exist for each character ID. 

One might be inclined to sample latent codes $w^{c}$ from the original latent space of $G_{swap}$ because it is a straightforward way of generating images with high fidelity for the StyleGAN2~\cite{stylegan2}. However, we found that it behaves differently for the layer-swapped generator (\Fref{fig:comp_g}). It is understood that the low-resolution layers of $G_{swap}$ produce feature maps for natural faces that are not favorable for high-resolution layers that are fine-tuned on the cartoon images. Therefore, we instead create $w^{c}$ explicitly by inverting the cartoon images and averaging them by each character ID in the $\mathcal{W+}$ space (Equation~\ref{eq:eq1}). With this operation, $w^c$ becomes more pose- and noise-invariant, which are desirable properties for the following style mixing procedure.
\smallskip

\noindent\textbf{Style mixing.}
Here, we define the style mixing step that blends the given input codes $w^{f}$ and $w^{c}$. To do that, we first pass them through the affine layer $A$ to obtain the corresponding style parameters $s^{f}, s^{c} \in \mathcal{S}$ (Equation \ref{eq:cdsm}).
Consequently, parts of $s^f$ are replaced by those of $s^c$ using the tRGB replacement method (discussed in \Sref{subsec:trgb_replacement}).
Two sets of style parameters, i.e., $s^f$ and $s^{c}$, are finally style mixed in the $\mathcal{S}$ space as follows:
\begin{equation}
SM(s^f, s^{c}, m) \equiv \{s^{f}_{1,\dots,t(m)-1}\} \cup \{s^{c}_{t(m),\dots,26}\}.
\end{equation}
where $t(m)$ is the mapping function that maps the indices of the latent code $w \in \mathcal{W+}$ to the corresponding style parameters $s \in \mathcal{S}$. 
% We choose to perform style mixing in the $\mathcal{S}$ space because it enables color-specific manipulation owing to its disentangled nature.

\figCompColor{fig:comp_color}

\subsection{tRGB replacement}
\label{subsec:trgb_replacement}

In our framework, the encoder for the source domain is bound to generate latent codes in the $\mathcal{W+}$ space of the layer-swapped generator. We hypothesize that utilizing this space can potentially cause color distortion since the original $\mathcal{W}$ space shows no sign of color artifacts at all (\Fref{fig:comp_color}b).
It is also suggested that one can change only the color of the local or global regions of an image by manipulating only parts of the style parameters ${s_{tRGB}}$ that control the \texttt{tRGB} layers in StyleGAN2, leaving all the semantic features intact~\cite{stylespace}. In this context, we found that the encoder produces ${s_{tRGB}}$ deemed as out-of-distribution for the \texttt{tRGB} layers, resulting in severe color distortion (\Fref{fig:comp_color}a). On this basis, we developed the \texttt{tRGB} replacement method to pin-down the cause of the color distortion to the style parameters and to manipulate only ${s_{tRGB}}$ to alter the color distribution of the output image. Specifically, the \texttt{tRGB} parts of the style parameters of the natural face image ${s^f_{tRGB}}$ are replaced by the counterparts of the target cartoon character ${s^c_{tRGB}}$, thus ensuring that ${s^f}$ possesses the \texttt{tRGB} parts representing the color distribution of the target character images (\Fref{fig:overview_framework}).
\begin{equation}
R(s^{f}, s^{c}) \equiv s^f \setminus \{s^f_{tRGB}\} \cup \{s^c_{tRGB}\}
\end{equation}
We show that this method not only corrects the color distortion issue (\Fref{fig:ablation}b), but also
helps with transferring character-specific fine features when applied along with the style mixing component (\Fref{fig:figtRGB}).

\figtRGB{fig:figtRGB}

% Later, we will show that, with this method in place, the color distortion issue is completely resolved (\Fref{fig:ablation}).

% In this context, we found that the encoder produces ${s_{tRGB}}$ deemed as out-of-distribution for the \texttt{tRGB} layers (\Fref{fig:comp_color}), resulting in the severe color distortion.

\figQual{fig:qual}
% \tableQuant{tab:quant}

\section{Experiments \& Applications}
\label{sec:experiments}

We first fine-tuned the FFHQ pretrained StyleGAN2 with ADA~\cite{ada} on the cartoon dataset using the default settings, except for the style mixing probability, which was set to zero. The fine-tuned generator was layer-swapped with the FFHQ pretrained generator at a $32\times32$ resolution. By default, the style mixing level $m$ was set to six. For each character ID, we randomly sampled $k=50$ images from the training dataset and inverted these into the $\mathcal{W+}$ space to prepare the character ID-specific latent codes $w^c$.

\subsection{Model comparison}
\noindent\textbf{Dataset.}
We constructed a cartoon dataset of 1226 images and 13 character IDs (each ID has $<$100 images).
We collected the cartoon characters from iCartoonFace~\cite{icartoon} and Webtoon, a type of digital cartoon platform.
Because we gathered characters from various cartoons, our dataset had diverse face abstraction (d\'eformer) levels, unlike the previous portrait stylization datasets.
All the images were super-resolved to 1024$\times$1024 using SRCNN~\cite{dong2015image}.
\smallskip

\noindent\textbf{Baselines.}
We compared our work with CycleGAN~\cite{cyclegan}, U-GAT-IT~\cite{ugatit}, Toonify~\cite{toonify}, and AgileGAN~\cite{agilegan}.
CycleGAN and U-GAT-IT are two representative unpaired I2I translation models, whereas Toonify and AgileGAN are StyleGAN-based methods.
Because none of the competing methods can handle multiple character IDs with a single network, we built separate models for each character.
CycleGAN and U-GAT-IT were trained on the FFHQ-aligned selfie dataset, which was resized to $256\times256$ owing to the GPU memory availability. For a fair comparison, we downsampled the outputs of Toonify, AgileGAN, and ours to $256\times256$ with Lanczos filtering.

\smallskip
\noindent\textbf{Model Comparison.}
The I2I-based methods (CycleGAN and U-GAT-IT) cannot preserve the global structure of the source image (\Fref{fig:qual}a-b). This is because the cycle consistency makes it difficult for them to translate the highly disparate geometric shapes between two domains (\textit{e.g.,} jaw).
In contrast, the StyleGAN-based methods (\Fref{fig:qual}c-e) have a better ability to maintain the identity features.
Among them, Toonify suffers from color distortion issues and is limited in expressing the fine details of the target character.
For AgileGAN, despite the exhaustive training resources, it cannot generate abstract cartoon faces adequately. We hypothesize that the similarity loss used in AgileGAN results the preservation of a large part of the natural face texture and shape.
Our model not only significantly resolves the color artifacts and preserves the global structure of the input image but also successfully expresses the fine details of the target character.
% \figColorDistort{fig:color_distort}
% \tableQuant{tab:quant}

We also present the results of the quantitative evaluations, including the user study. In this benchmark, we used 289 cartoon images with three randomly selected  character IDs as the training dataset and randomly sampled 781 face images from the FFHQ validation dataset as the test dataset. FID~\cite{fid}
% Fr\'echet inception distance
% kernel inception distance  
and KID~\cite{kid} were used to evaluate the stylized results of the test dataset. Regarding the user study, we quantified the score by averaging the preference rates of the 40 professionals in the cartoon industry. Each participant was shown 10 randomly selected face images along with their stylized results. As shown in \Tref{tab:quant}, one notable observation is that the I2I-based methods achieved the lowest FID and KID despite the inferior visual quality and user preference score. We argue that the limited number of the training images ($<$100 images for each cartoon character) makes the evaluation measure unreliable. In addition, a highly abstract face texture in the cartoon domain may not be suitable for conventional metrics in this category.
In the StyleGAN-based category, our method outperformed the others in all the measurements by a large margin, including the user preference score. We would also like to emphasize that all the results of our method were from a single generator by simply switching the ready-to-use character latent codes. In contrast, other methods should build $N$ networks to stylize the faces into $N$ characters.

\tableQuant{tab:quant}

\figAblation{fig:ablation}
\figCompLatentspace{fig:comp_latentspace}

% \figAblation{fig:ablation}
% \figCompLatentspace{fig:comp_latentspace}
\subsection{Internal Analysis}
\noindent\textbf{Ablation study.}
We analyzed how each component of our method improved the quality and stability of the produced image.
The color distortion issue was completely resolved by the \texttt{tRGB} replacement component (\Fref{fig:ablation}b). The style mixing in StyleSpace component effectively expresses the essential style features of the target character (\Fref{fig:ablation}c).
By combining both components, our full model can perform a decent quality of stylization without any quality degradation (\Fref{fig:ablation}d).
In contrast, the conventional layer swapping method severely suffers from color artifacts and lack of features of the target character (\Fref{fig:ablation}a).
\smallskip

\figColorDistort{fig:color_distort}

\noindent\textbf{Choice of the latent space.} We demonstrate how the choice of latent space affects the preservation of the cartoon features discussed in \Sref{subsec:stylemixing}. Here, we inverted the input image to the desired latent space and decoded it with our layer-swapped generator.
With the latent code in the $\mathcal{W}$ space, the generated image tends to be blurry and, more importantly, it lacks detailed cartoon features such as the unique mouth shape and eyelashes (\Fref{fig:comp_latentspace}b).
In contrast, the $\mathcal{W+}$ space can preserve the fine features in great detail (\Fref{fig:comp_latentspace}c).
These results show the superior expressiveness of the $\mathcal{W+}$ space even for the cartoon domain, justifying the use of $\mathcal{W+}$ for inversion by default in our framework.
\smallskip

\noindent\textbf{Color distortion.} We found that only the \texttt{tRGB}-treated style parameters were linked to the aforementioned color distortion issue in our work. Like in the experiment conducted in StyleSpace~\cite{stylespace}, we deliberately perturbed only the \texttt{tRGB} parts of the style parameters to determine their impact on the output image as follows: 
\begin{equation}
{s_{{tRGB}_{perturbed}}} = {s_{tRGB}} + N *{s_{tRGB}}.
\label{eq:color_pertubation}
\end{equation}
where $N$ is the manipulation strength. As shown in \Fref{fig:color_distort}, the perturbed tRGB parts only affect the color distribution, leaving the other features intact.

% \figApplication{fig:application}
% \figConsistency{fig:consistency}
% \figFailure{fig:failure}

\subsection{Applications}
In this section, we present a downstream application using our Cross-domain Style mixing method as a core module. One possible scenario is a photo-to-cartoon (\Fref{fig:application}b), which provides an immersive experience wherein a user becomes part of his/her favorite cartoon with the whole photo stylized, not just the facial or torso parts. Several studies on cartoonization~\cite{cartoongan,wbox} tackled a similar scenario; however, they could not express the essential features of the target character because the model was trained in an unpaired manner (\Fref{fig:application}a).
Alternatively, we performed the photo-to-cartoon conversion using the following pipeline: 1) face region detection, 2) face stylization with our proposed method, 3) scene translation via the cartoonization model, and 4) face and scene harmonization. Note that, with this pipeline, one can also build a video-to-cartoon application (\Fref{fig:consistency}), where the consistency of styles across video frames is critical, owing to the superior stability and robustness of our method.

\figApplication{fig:application}
\figConsistency{fig:consistency}
\figFailure{fig:failure}

\section{Conclusion}
\label{sec:conclusion}
In this paper, we presented a simple but strong portrait stylization method.
We first analyzed two drawbacks of the previous layer swapping method and mitigated these using our novel Cross-domain Style mixing method.
By doing so, our model 1) successfully resolves the color distortion and degenerated style diversity problems, and 2) produces superior stylization results using only a single generator for the cartoon characters at different face abstraction levels. Furthermore, we showed that our method can be extended to photo- or video-to-cartoon applications owing to its practicality and robustness.  
In the future, we will focus on how to properly stylize features not present in the target domain distribution (\textit{e.g.,} sunglasses and hats in \Fref{fig:failure}). We believe that this could be a fundamental problem from which many encoding-based translation methods suffer; however, mitigating this would help bring more robustness and reliability to real-world application scenario.

\balance

{
\small
\bibliographystyle{ieee_fullname}
\bibliography{}
}
\end{document}

\title{Supplementary Materials for \\``Cross-Domain Style Mixing for Face Cartoonization"}

\author{
Seungkwon Kim \qquad Chaeheon Gwak \qquad Dohyun Kim \qquad Kwangho Lee \\ Jihye Back \qquad Namhyuk Ahn* \qquad Daesik Kim* \\ NAVER WEBTOON AI
}

\twocolumn[{
\renewcommand\twocolumn[1][]{#1}
\maketitle
\appendix
\begin{center}
\centering
% \captionsetup{type=figure}
% \includegraphics[width=\linewidth]{figure/suppl/style_mixing/style_mixing.PNG}
% \caption{Stylized results with various style strength levels.}
% \label{fig:figSM}
\end{center}
}]

% \section{Experiments}
% \subsection{Effects of the tRGB Replacement Method} We show that the tRGB replacement method (Section 4.3) not only corrects the color distortion issue, but also helps with transferring character-specific style features.

% \subsection{Style Strength} 
% As shown in \Fref{fig:figSM}, one can determine the style strength of a target character in the output image by simply adjusting the style mixing level $m$ (Section 4.1). Specifically, bringing styles of a character-specific latent code ${w^c}$ from middle resolutions, (\textit{e.g.,} six; 16$\times$16) results in highest stylization quality in our case. This also shows that the conventional style categories (coarse, middle and fine styles) are still valid in the cross-domain setting.

% \figtRGB{fig:figtRGB}
\section{Additional Results}
\noindent\textbf{Face cartoonization.} Additional cartoonization results are presented in \Fref{fig:figResults}.
\smallskip

\noindent\textbf{Applications.} Additional examples of photo- and video-to-cartoon applications are presented in \Fref{fig:figPhoto} and \ref{fig:figVideo}.

\figResults{fig:figResults}
\figPhoto{fig:figPhoto}
\figVideo{fig:figVideo}